\def\year{2021}\relax
\documentclass[letterpaper]{article} 
\usepackage{aaai21}  
\usepackage{times}  
\usepackage{helvet} 
\usepackage{courier}  
\usepackage[hyphens]{url}  
\usepackage{graphicx} 
\usepackage{natbib}  
\usepackage{caption} 
\usepackage{multirow}
\usepackage{booktabs} 

\usepackage{amsmath} 
\title{DeepDT: Learning Geometry From Delaunay Triangulation \\ for Surface Reconstruction}
\author{
	Yiming Luo \footnotemark[2], Zhenxing Mi \footnotemark[2], Wenbing Tao\textsuperscript{}\footnotemark[1]\\

}
\affiliations{
    National Key Laboratory of Science and Technology on Multi-spectral Information Processing\\
    School of Artifical Intelligence and Automation, Huazhong University of Science and Technology, China \\
    yiming\_luo@163.com, mizhenxing.henan@gmail.com, wenbingtao@hust.edu.cn

}

\begin{document}

\maketitle

\renewcommand{\thefootnote}{\fnsymbol{footnote}}

\footnotetext[2]{Equal contribution.}
\footnotetext[1]{Corresponding author.}

\renewcommand{\thefootnote}{\arabic{footnote}}

\begin{abstract}
In this paper, a novel learning-based network, 
named DeepDT, is proposed to reconstruct the surface
from Delaunay triangulation of point cloud. DeepDT learns to predict inside/outside labels
of Delaunay tetrahedrons directly from a point cloud
and corresponding Delaunay triangulation. 
The local geometry features are first extracted from 
the input point cloud and aggregated into a graph 
deriving from the Delaunay triangulation.
Then a graph filtering is applied on the aggregated 
features in order to add structural regularization 
to the label prediction of tetrahedrons.
Due to the complicated spatial relations between tetrahedrons 
and the triangles, it is impossible to directly 
generate ground truth labels of tetrahedrons from 
ground truth surface. Therefore, we propose a multi-label supervision strategy which votes for the label of a tetrahedron with labels of sampling locations inside it.
The proposed DeepDT can maintain abundant geometry details without generating overly complex surfaces
, especially for inner 
surfaces of open scenes. Meanwhile, the generalization ability and time consumption of the proposed method is acceptable and competitive 
compared with the state-of-the-art 
methods. 
Experiments demonstrate the superior 
performance of the proposed DeepDT. 
\end{abstract}

\section{Introduction}

Surface reconstruction from 3D 
point clouds 
is a long-standing problem in 
computer vision and graphics \cite{ kazhdan2013screened}. 
A lot of previous methods use an implicit function
framework \cite{curless1996volumetric, kazhdan2006poisson, kazhdan2013screened}. They typically discretize space in the bounding box
of the input point cloud with a voxel grid or an adaptive octree.
Then they compute an implicit function from input points.  
The result surface is extracted from the grids as an 
isosurface of the implicit function by Marching Cubes (MC) 
\cite{lorensen1987marching}.
However, solving equations of large scale in implicit methods can be time-consuming. The maximum resolution of the octree depth also influences the efficiency. There is a quadratic relation between resolution and run time as well as memory usage. Post-processing steps are also needed to clean up the excess part in result surfaces sometimes (trimming), which are sensitive to corresponding parameters and make batch processing of point clouds impractical.
This family of implicit approaches is sometimes limited by their sensitivity to noise, outliers, non-uniform sampling or even simply by the lack of reliable and consistent normal estimates and orientation \cite{labatut2009robust}. 
Another common method 
uses a Delaunay
triangulation of a point cloud to subdivide 
the space into uneven tetrahedrons. 
As analyzed in \cite{amenta1999surface}, 
under the assumption of sufficiently dense point clouds,
a good approximation
of the surface is contained in the Delaunay
triangulation. Therefore, opposed to
the approaches that try to fit a surface 
in a continuous space, surface reconstruction 
based on Delaunay triangulation can now be reduced 
to selecting an adequate subset of 
the triangular facets in the Delaunay
triangulation.
\begin{figure}
	\begin{center}
		\includegraphics[width=\linewidth]{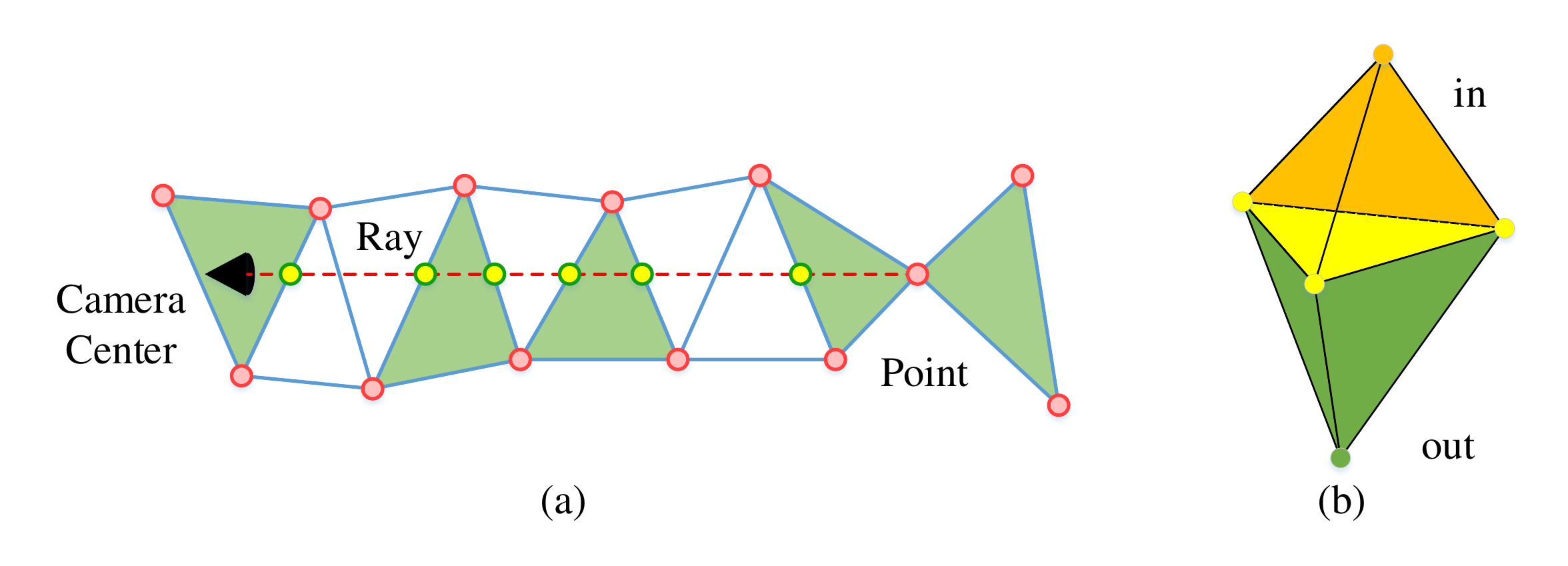}
	\end{center}
	\caption{A 2D example of reconstructing
		surface by in/out labeling of tetrahedrons.
		The visibility information is 
		integrated into each tetrahedron
		by intersections between viewing rays and
		tetrahedrons. 
		A graph cuts optimization
		is applied to classify tetrahedrons
		as inside or outside the surface.
		Result surface is reconstructed by
		extracting triangular facets between 
		tetrahedrons of different labels. }
	\label{fig:intro}
\end{figure}
Researchers have spent a lot of efforts 
in finding the object surface 
in a Delaunay triangulation.
A commonly used strategy 
is to classify the tetrahedrons as 
inside/outside the surface using graph cuts. 
Visibility information is required 
in such a process. A 2D example is shown in Figure
\ref{fig:intro}. 
Such method builds a directed graph based on the neighborhood of adjacent tetrahedrons in the Delaunay triangulation. Visibility terms make this kind of approaches more accurate and robust to outliers.

However, such method still faces problems. Some tetrahedrons
lying behind points risk being mislabeled
because 
no viewing rays from 
the camera center to the points will ever 
intersect them. 
Therefore, such method could
generate overly complex 
bumpy surfaces with
many unexpected handles inside the model, failing to get clean inner surfaces.
What's more, such a method fails to cope with
arbitrary point clouds without visibility information. 
It suggests that correct labeling of tetrahedrons without visibility information is still a bottleneck to overcome.

Most recently, the rapid development 
of deep learning  
and improvement of large-scale 3D datasets 
make it possible to train neural networks for 3D shapes. 
A variety of learning-based surface reconstruction methods
have been proposed, such as learning to refine 
Truncated Signed Distance Function (TSDF)
on voxel by 
voxel networks or octree grids by octree networks 
\cite{dai2017shape, riegler2017octnetfusion, cao2018learning},
learning occupancy and SDF functions from point clouds
\cite{mescheder2019occupancy,park2019deepsdf}. 
Learning-based methods are inclined to introduce more 
geometry priors into surface reconstruction 
for better performance. However, existing 
learning-based methods still face 
problems: 
(1) Most of them still rely on voxel or octree
structures, which are not prone to maintain high computational efficiency. 
(2) Their ability to generate ﬁne-grained structures and details is still 
limited due to the low-resolution girds 
or the latent vector of a fixed size.
(3) Most of them still rely on global features of training data excessively, which leads to generalization problems.

Another excellent work, SSRNet \cite{Mi_2020_CVPR}, the state-of-the-art learning-based method in Surface Reconstruction from Point Clouds (SRPC) task to our known, constructs local geometry-aware features, which leads to accurate classification for octree vertices and outstanding generalization capability among different datasets. However, it is worth noting that not all inputs are dense enough and there may be missing data in the input.

This inspires 
us to explore a better learning-based 
method, which is able to finish reconstruction effectively and efficiently without visibility information.
In this paper, we propose DeepDT, a novel learning-based method 
for surface reconstruction from 
point clouds based on Delaunay triangulation. 
We construct Delaunay triangulation from the point cloud
 and replace 
graph cuts with a network
for accurate labeling of tetrahedrons.
The network takes a point cloud and its Delaunay 
triangulation as input, and learns to classify the 
tetrahedrons as inside/outside.
In our method, the key insight is that the inside/outside labeling
of tetrahedrons can be determined via local geometry 
information of points and the structural 
information of Delaunay triangulation.
We first encode features of points with the geometry feature extraction
module. It exploits local K nearest neighbors
to encode a local geometry feature for each point.
Meanwhile, it directly encodes features representing in/out 
information, such as signed distances. 
Therefore, it is able to capture local geometry details 
of the surface. 
We then augment the graph of Delaunay 
triangulation with geometry features by a point-to-graph
feature aggregation operation. We use an attention mechanism to aggregate geometry features into 
graph nodes. It can automatically 
select important geometry features for in/out 
classification of tetrahedrons. 
Subsequently, we apply a graph filtering module 
to the feature augmented graph. The module 
exchanges information among neighboring tetrahedrons
with Graph Convolution Networks (GCN) \cite{kipf2016semi}  
and thus adds structural constraints to the label prediction.
Our network is trained in a supervised manner. 
However, it brings about a problem that we cannot directly get the ground 
truth labels of tetrahedrons from ground truth 
surfaces. 
Since tetrahedrons are 
likely to have complicated intersections with 
triangular surfaces, we cannot obtain the relative 
position relationship between the tetrahedron 
and the triangular mesh. To tackle this problem, we 
propose a multi-label supervision strategy. 
We obtain the in/out labels of tetrahedrons through a multi-label supervision strategy which  utilizes the labels of multiple reference locations sampled inside them.
As analyzed above, our method 
enjoys a combination of novelty as follows. 
\begin{itemize}
	\item Our method integrates geometry and graph structural 
	information for more robust tetrahedron labeling.
	\item Local geometry features together with 
	global graph structural regularization 
	benefits our method with good accuracy and generalization
	capability.
	\item The multi-label supervision mechanism makes it possible 
	to train a high quality model without 
	ground truth labels of tetrahedrons or visibility information.
	
\end{itemize}

Experiments on challenging data show that our method can complete the reconstruction accurately and efficiently, and restore the geometric details of the input data with noise and complex topology.
It is also proved that our method can generalize 
well across datasets of different styles.

\section{Related Work}

\paragraph{Geometric reconstruction methods}

Implicit reconstruction methods \cite{levin2004mesh, guennebaud2007algebraic, fuhrmann2014floating, kazhdan2013screened, curless1996volumetric, carr2001reconstruction, turk2002modelling}
attempt to approximate the surface as 
an implicit function from which 
an isosurface is extracted by Marching Cubes (MC) \cite{lorensen1987marching}. 
\cite{hoppe1992surface}
estimates a tangent plane for each point with k-nearest
neighbors. The implicit function is defined as the signed
distance to the tangent plane of the closed point.
Poisson surface reconstruction method (PSR)
\cite{kazhdan2006poisson} 
fits the vector field 
of point normals with gradient 
of an in/out indicator function.
Since the discretization is regular, these methods 
are mostly suitable for compact point clouds with tight bounding boxes.
Further more, a post-processing step is needed to obtain an explicit surface from an implicit function, known as ray tracing and MC.

Another common strategy to complete reconstruction task works with Delaunay tetrahedralization of the input points instead of voxel-based volumetric representation \cite{vu2011high, jancosek2011multi, hoppe2013incremental, hiep2009towards}.
\cite{labatut2007efficient} is the first to 
propose a 
reconstruction method (L-Method) that classifies tetrahedrons 
as inside/outside by visibility information and
graph cuts. 
Many later methods mostly focus on
adding different terms to the graph cuts 
 in order to improve the 
surface quality \cite{labatut2009robust, jancosek2014exploiting, zhou2019detail}. 
The strategy of visibility constraints 
and graph optimization has achieved 
remarkable performance on challenging data with noise and outliers. However,
it fails to cope with arbitrary
point clouds without visibility information. 
Even with visibility information, 
it is usually not enough 
to label tetrahedrons behind a point correctly.

\noindent{\textbf{Learning-based methods  }} 
Recently developing deep learning 
techniques have given rise to
lots of learning-based reconstruction methods \cite{liao2018deep, groueix2018papier}. 
The networks based on voxel or octree
grids \cite{riegler2017octnetfusion, dai2017shape, riegler2017octnet, cao2018learning} usually face efficiency issues
so they can only reconstruct surfaces 
of compact objects at a relatively low resolution.
Instead of optimizing TSDF
on grids, some learning-based
methods directly learn a continuous function
from point clouds.
The ONet
encodes the point cloud into a global latent vector.
It predicts occupancy values for 
3D locations by decoding the latent vector
concatenated with 3D coordinates. 
DeepSDF \cite{park2019deepsdf} is similar to 
ONet, but it predicts signed distance values. The 
result surface is also extracted by MC
from an octree.
Learning-based methods 
using voxel or octree grids to discretize
the surface are sharing similar problems with the traditional implicit methods.
Moreover, most existing learning-based methods encode structural details into fixed-size latent vectors. Although they can usually generate a final surface that roughly retains the shape of the input, they may not be able to capture geometric details of complex topologies.
In addition, most existing methods learn too many global features, which makes them perform well on categories similar to the training set. However, they usually have difficulty maintaining the generalization ability well across different datasets.

\section{Method}

In this section, we describe the detailed network architecture
of DeepDT. Figure \ref{fig:pipeline}(a) is a 2D illustration 
of our main pipeline.
The input of our network includes a point cloud 
$\mathcal{P}$ (black dots in the left of Figure \ref{fig:pipeline}(a)) 
and its Delaunay triangulation $\mathcal{D}$ 
(black triangles in the left of Figure \ref{fig:pipeline}(a)). 
$\mathcal{P}$ is a set 
of 3D points with their normals. $\mathcal{D}$
is a set of tetrahedrons.
The four vertices of each tetrahedron
are points in the
$\mathcal{P}$. Each tetrahedron also has four neighbor tetrahedrons
sharing common triangular facets. 
The Delaunay triangulation
structure forms a graph  $\mathcal{G}$ 
(red dots connected by red lines in the left of 
Figure \ref{fig:pipeline}(a)), with tetrahedrons as nodes
and common triangular facets connecting two adjacent
tetrahedrons as edges. 
Our network can be divided into two parts. 
One is mainly about geometry feature extraction.
The other is for graph feature aggregation and filtering.

\subsection{Geometry Feature Extraction} \label{sec:Geometry_Feature_Extraction}

The point features in our method provide initial
geometry information for in/out classification of 
tetrahedrons. Therefore, the layer of geometry feature extraction module should have three 
properties to better complete local surface feature encoding.
(1) It must directly provide in/out information 
with respect to the implicit
surface.
(2) It should encode in/out information locally
in order to capture the geometry details.
(3) It has to be computationally efficient in order
to process larger point clouds. 
The computational efficiency and local encoding are 
related to the network architecture while the 
in/out information is determined by the input features.

\begin{figure*}
	\begin{center}
		\includegraphics[width=0.85\linewidth]{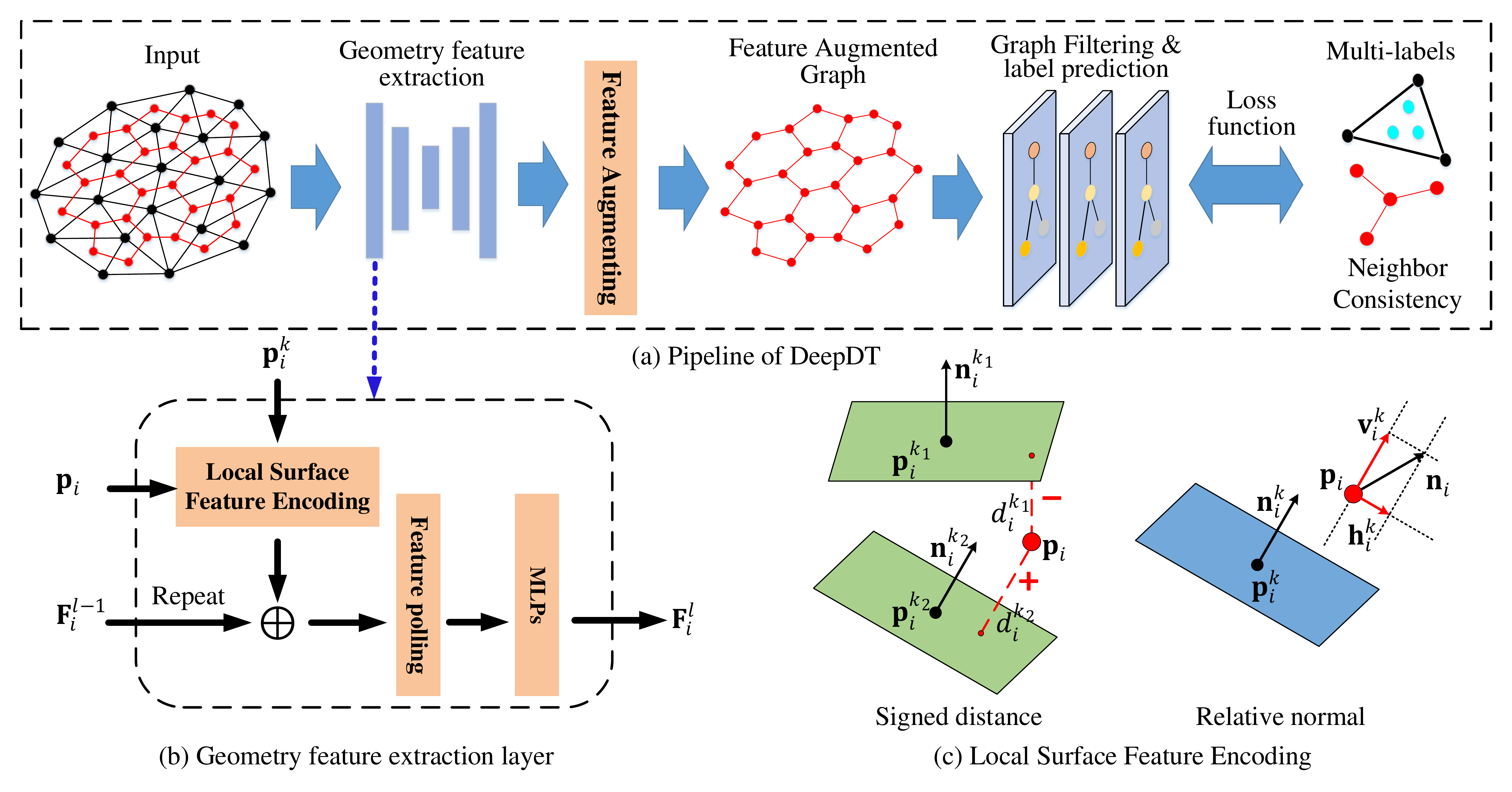}
	\end{center}
	\caption{(a) The pipeline of our method. 
		The input includes a point cloud (black dots) 
		and its Delaunay triangulation
		(black triangles). 
		The Delaunay triangulation
		structure forms a graph
		(red dots connected by red lines).
		(b) Detailed illustration of 
		geometry feature extraction layer, $ \oplus $ means concatenation. (c) Calculation 
		of signed distances and relative normals.}
	\label{fig:pipeline}
\end{figure*}

\noindent{\textbf{Local Surface Feature Encoding  }} 
  We compute signed distances among each point 
and its neighbors as the raw input feature of our network. 
For each reference point $ \mathbf{p}_i $, we search 
its $ K $ nearest neighbors $ \mathbf{p}_{i}^k, k=1,...,K $.
The neighbor $ \mathbf{p}_{i}^k $ with 
normal $ \mathbf{n}_{i}^k $ can be seen as a 
tangent plane $ t_{i}^k $ approximating the local surface 
near $ \mathbf{p}_{i}^k $. Note that normals of points are all
normalized so their lengths are equal to 1.
As illustrated in Figure \ref{fig:pipeline}c, signed distance
$ d_{i}^k $
between  $ \mathbf{p}_i $ and tangent plane can be calculated
as:
\begin{equation}
	d_{i}^k = (\mathbf{p}_i - \mathbf{p}_{i}^k) \cdot \mathbf{n}_{i}^k, \quad \| \mathbf{n}_{i}^k \| = 1
\end{equation}
The symbol $ \cdot $ means dot product of vectors.

In addition to the signed distances, we also include relative 
normals in the input feature in order to provide 
richer information about local surface geometry.
For each reference point $ \mathbf{p}_i $ 
with normal $ \mathbf{n}_{i} $ and one of its 
neighbor tangent plane $ t_{i}^k $,
we decompose
$ \mathbf{n}_{i} $ into two vectors 
$ \mathbf{v}_{i}^k $ and $ \mathbf{h}_{i}^k $.
, relative to the 
tangent plane. 
$ \mathbf{v}_{i}^k $ is perpendicular to $ t_{i}^k $
and $ \mathbf{h}_{i}^k $ is parallel to $ \mathbf{t}_{i}^k $.
They can be calculated as:
\begin{equation}
	 \mathbf{v}_{i}^k = (\mathbf{n}_{i} \cdot \mathbf{n}_{i}^k) \mathbf{n}_{i}^k, \;
	 \mathbf{h}_{i}^k = \mathbf{n}_{i} - \mathbf{v}_{i}^k, \; \| \mathbf{n}_{i} \| = \| \mathbf{n}_{i}^k \| = 1 
\end{equation}
The symbol $ \cdot $ means dot product of vectors.
The calculation is illustrated in Figure \ref{fig:pipeline}c.
After computing the signed distances and relative normals,
the surface feature $ s_{i}^k $ between $ \mathbf{p}_i $ and 
$ \mathbf{p}_{i}^k $ is encoded as:
\begin{equation}
	s_{i}^k = \text{MLP}(d_{i}^k \oplus \mathbf{v}_{i}^k \oplus \mathbf{h}_{i}^k)
\end{equation}
$ \text{MLP} $ means Multi-Layer Perceptron. 
$ \oplus $ means concatenation.
By directly encoding the local surface features,
the point features can provide in/out information for
the classification of tetrahedrons. 
As a reward for using only local 
geometry features rather than global
features, our network 
generalizes well across different datasets.

After designing Local Surface Feature Encoding, 
we have completed the most important part of geometric feature extraction.
In Figure \ref{fig:pipeline}a, the geometry feature extraction
network consists of multiple geometry feature extraction layers
shown in Figure \ref{fig:pipeline}b. Besides, random downsampling 
expands the receptive fields. The $ l $-th multiple geometry 
feature extraction layer
takes the reference point $ \mathbf{p}_i $, the previous feature 
$ F_i^{l-1} $ and $ K $ nearest neighbors 
$ \mathbf{p}_{i}^k, k=1,...,K $ as input. It first encodes 
local surface features for $ \mathbf{p}_i $ with respect to
each neighbor point. Then $ F_i^{l-1} $ is repeated K times and 
concatenated with K local surface features. After that, the K
features are aggregated by an attention pooling layer, which aims to use the attention mechanism to help automatically learn important local features.
A group of MLP layers are applied to the aggregated
feature and output result feature $ F_i^{l} $ as 
$ l $-th feature of $ \mathbf{p}_i $.

\subsection{Feature Augmented Graph}\label{sec:Feature_Augmented_Graph}
\noindent{\textbf{Point to Graph Feature Aggregation  }} 
After extracting the features for each point, 
we build a feature augmented graph from the point features and
the graph $\mathcal{G}$.
The graph directly derives from the Delaunay triangulation $\mathcal{D}$ of the 
point cloud. 
The graph nodes and graph edges correspond to Delaunay tetrahedron and the triangular facets between adjacent tetrahedrons.
Each tetrahedron in $\mathcal{D}$ consists of 
four vertices from the point cloud. 
Therefore, we construct the tetrahedron features in the graph 
by aggregating the geometry features of the four vertices. 
Let $ T_{i} $ be the feature of $ i $th node in  $\mathcal{G}$ with its four point
features $ F^{j}_{i} (j=1,2,3,4) $.
We construct $ T_{i} $ from the four vertex 
features by an attention mechanism, which is able to channel-wisely select 
geometry features important for the classification of tetrahedrons.
We first learn weights for each channel of the four features.
Let $ F_{i}=\{F^{j}_{i}, j=1,2,3,4\} $ 
be the set of four features with shape $ (4, C) $. 
$ W_{i} $ is the set of four weight 
vectors with the same shape $ (4, C) $. 
$ C $ means the number of channels. The Softmax operation is applied to $ W_{i} $
in its first dimension. Then corresponding channels are weighted averaged.
\begin{equation}
	W_{i} = \text{Softmax}(\text{MLP}(F_{i}))
\end{equation}

\begin{equation}
	T_{i} = \sum W_{i} \odot F_{i}
\end{equation} 
$ \odot $ is the element-wise product of $ W_{i} $ and $ F_{i} $.
$ \sum $ is applied to the first dimension.
$ T_{i} $ has the shape of $ (1, C) $.

A special case is that the Delaunay triangulation has
some infinite tetrahedrons which have three
vertices in the 3D point cloud and share
an infinite vertex. To solve this problem,
we directly set the feature of the infinite vertex
as zeros.

\noindent{\textbf{Graph Filtering  }} 
The raw tetrahedron features in the graph $\mathcal{G}$ constructed 
by aggregating geometry features
encode the inside/outside information of tetrahedrons.
They are constructed from point features
independently so they do not contain enough neighborhood graph information.
They could contain 
noise and inconsistency due to the 
existence of noise in the point clouds. 
The classical graph cuts based methods have demonstrated
that local neighborhood smoothing constraint 
in a graph
is important for
robust label prediction for graph nodes.
Therefore, after the 
construction of graph features,
we apply the multi-layer Graph Convolutional
Network
to the graph in order to 
integrate more local graph structural constraints. 
The GCN layers refine the tetrahedron features
by exchanging information
among neighboring tetrahedrons. 
It is able to encode more graph structural information 
for label prediction.
The last layer of the GCNs outputs a 
2-channel prediction vector for each tetrahedron.
The softmax operation is finally applied to the vectors
in order to get the probability for a tetrahedron to be inside/outside.

\subsection{Loss Functions}

\noindent{\textbf{Multi-label Supervision  }}
A straightforward supervision 
for our network is the ground truth labels
of the tetrahedrons. 
However, 
due to the very complicated spatial relations the tetrahedrons
have with the triangles, it is not trivial to label the
tetrahedrons via the ground truth triangular surface.
The tetrahedrons may have intersections with multiple triangles
so the classification is ambiguous. One compromise approach
is to use classical graph cuts based methods for label generation. 
However, this will limit the accuracy potential of the network.
In our method, we propose a multi-label supervision method. Since
it is easy to check whether a 3D location is inside/outside a surface,
we randomly sample $ N $ reference locations ($ N\_ref $) in 
a tetrahedron and get their inside/outside 
labels. Then we use these labels to 
supervise the labeling process of the tetrahedron.
Each label gives a “vote” for each tetrahedron
to determine in/out label.
With the aforementioned architecture and multiple “votes”, 
our method is able to encode 
more accurate relations among the 
tetrahedron labels, the point geometry and 
Delaunay triangulation structure. 
It is worth noting that these reference locations
only provide labels for each tetrahedron. The locations
themselves are not used in our network.

To train our network with multi-label supervision,
we compute a multi-label loss $ L_{m} $. 
It is a classification loss minimizing the 
error between predicted probabilities 
and multi-labels. 
With a Delaunay triangulation
of $ N $ tetrahedrons, we sample $ N\_ref $ reference locations
for each tetrahedron $ T_{i} $ and get $ N\_ref $ labels.
The $ f_{ij} $ is the binary cross-entropy loss 
between the prediction probabilities of $ T_{i} $
and its $ j $th label $ l_{i}^{j}. $
$ L_{m} $ gives more help to label predication with $ N\_ref $ rising. This will be very beneficial to the prediction of very large tetrahedrons. It  can be denoted by:
\begin{equation}
	L_{m} = \dfrac{1}{N \times N\_ref}\sum_{i=1}^{N}\sum_{j=1}^{N\_ref}f_{ij}
\end{equation}

\noindent{\textbf{Neighbor Consistency Constraint  }}
In classical graph cuts based methods, the label smoothness
of neighbor tetrahedron  
helps to
reconstruct more smooth surfaces.
In our method, we
explicitly introduce a regularization 
loss for more consistent labeling among adjacent 
tetrahedrons. 
This loss encourages more smooth surfaces 
and also helps to classify the 
very large tetrahedrons connecting
points far from each other. 
With a Delaunay triangulation
of $ N $ tetrahedrons, the $ i $th tetrahedron $ T_{i} $ has four 
neighbor tetrahedrons 
$ T^{j}_{i},(j=1,2,3,4) $. 
The $ g_{ij} $ is the cross-entropy between 
the prediction probabilities between 
$ T_{i} $ and its $ j $th neighbor $ T^{j}_{i} $.
Then the multi-label loss in computed as:
\begin{equation}
	L_{n} = \dfrac{1}{4N}\sum_{i=1}^{N}\sum_{j=1}^{4}g_{ij}
\end{equation}
With $ \lambda_{1}$, $ \lambda_{2}$ balancing the two 
losses, our loss function can be summarized as:
\begin{equation}
	L = \lambda_{1} L_{m} + \lambda_{2} L_{N}
\end{equation}

\subsection{Surface Extraction And Data Preparation}

After getting the label of each tetrahedron, 
the output triangular surface is generated
by taking the triangular facets between adjacent
tetrahedrons with different labels.
We then post-process the result surface by Laplacian-based 
smoothing method for a more smooth surface and better
visual effects.
We prepare training data from a point cloud and its ground truth surface.
First, Delaunay triangulation is constructed from the point cloud using
the Computational Geometry Algorithms Library 
(CGAL) \cite{boissonnat2000triangulations}. The point indices for 
each tetrahedron and the adjacent matrix used for graph convolution 
are derived trivially from the Delaunay structure.
Then $ N\_ref $ locations are randomly sampled in each tetrahedrons.
We compute the inside/outside labels of these locations with 
respect to the ground truth surface and record these location labels.

\begin{table*}[t]
	\centering
	{
		
		\begin{tabular}{lccccccr}
			\toprule
			\multicolumn{1}{l}{Metric\textbackslash{}Method}& DMC   & ONet  & LDIF  & CEISR & SSRNet & Ours  &  \\
			\midrule
			
			Chamfer-$ L_1 $   & 1.17  & 0.79  & 0.40   & 0.41  & 0.24  & \textbf{0.20} &  \\
			NC          & 0.848 & 0.895 &  --     & 0.902 & \textbf{0.967} & \textbf{0.967} &  \\
			\bottomrule
	\end{tabular}}
	\caption{Quantitative results on ShapeNet. We evaluate methods with the Chamfer-L1 distance (lower is better) and Normal Consistency (higher is better).}
	\label{tab:results_on_shapenet}
	
\end{table*}

\section{Experiments}
In this part, we perform a series of experiments on datasets of different scales to qualitatively and quantitatively evaluate our DeepDT from different perspectives.

\noindent{\textbf{Datasets \& Evaluation Metrics  }}
We select three datasets with quite different types to comprehensively compare our DeepDT with traditional methods and other state-of-the-art learning-based methods. They are ShapeNet \cite{chang2015shapenet}, DTU \cite{jensen2014large} and Stanford 3D. \footnote{http://graphics.stanford.edu/data/3Dscanrep/} 
We choose the Chamfer-$ L_1 $ distance and the normal consistency (NC) score for experiments on ShapeNet.
For DTU dataset, we follow the DTU Completeness metric given by DTU and also take Chamfer Distances (CD) into consideration. We compute the CD between the ground truth point cloud and the vertices of the result surface that is expressed by the triangular meshes. For each point in a cloud, CD finds the nearest point in the other point set, and averages the square of distances up.
When it comes to Stanford 3D, we evaluate Chamfer Distances in generalization capability tests. In the following experiments, unless otherwise specified, the number of reference locations ($ N\_ref $) is set to 5, and the $ \lambda_{1}$, $ \lambda_{2}$ are set to 0.9 and 0.1, respectively.

\begin{figure}
	\begin{center}
		\includegraphics[width=\linewidth]{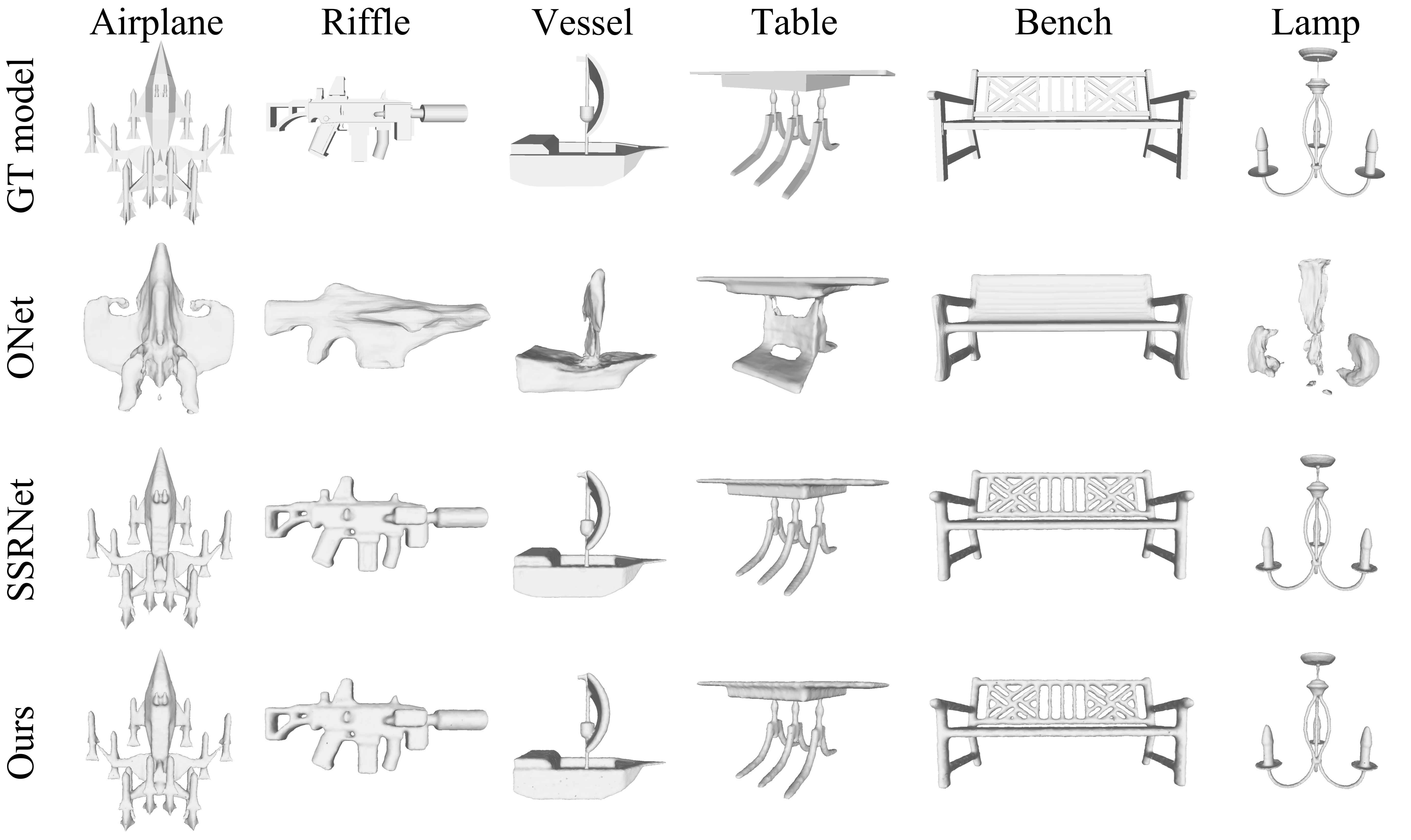}
	\end{center}
	\caption{Qualitative results of ShapeNet test data.}
	\label{fig:test1}
\end{figure}

\subsection{Results on ShapeNet}\label{sec:resultsonshapenet}
In this section, we compare our method with several state-of-the-art learning-based methods on ShapeNet : DMC \cite{liao2018deep}, ONet, LDIF \cite{Genova_2020_CVPR}, CEISR \cite{Poursaeed20a}, SSRNet. For fair comparison, we adopt the same train/validation/test split (about $ 30K $/$ 4.5K $/$ 9K $ shapes) as the mentioned methods.
Since data in ShapeNet is synthetic, we apply noise using a Gaussian distribution with zero mean and standard deviation 0.05 to the point clouds as ONet and SSRNet did. In order to get the labels of sampled reference locations, we reconstruct surfaces through PSR (octrees depth=$ 9 $) to generate training data. 
Quantitative evaluation results for meshes generated by DeepDT are reported in Table \ref{tab:results_on_shapenet}. We can find that the NC score of our method is comparable to that of the state-of-the-art learning-based method in SRPC task, SSRNet, and our method performs the best among all mentioned methods with the evaluation metric of Chamfer-$ L_1 $.

Figure \ref{fig:test1} shows that our method outperforms ONet in recovering shape details, especially highlighting the ability to retain as many details as possible on extremely complex objects. We can find that ONet can still barely cope with simple topologies well, but its performance on complex topologies is hard to satisfy us who expect it to perform better. 
This is mainly because the use of the global latent vector encoder makes ONet lose a lot of geometric details, but geometric details are essential for reconstructing complex topologies. This also reminds us that most learning-based methods that encode the entire shape using a global latent vector are facing a non-negligible problem, that is, the geometry details can be captured by the feature space which is limited by the ﬁxed size of the latent feature vector. We cannot solve this problem simply from the perspective of downsampling the input.

\begin{figure}
	\begin{center}
		\includegraphics[width=\linewidth]{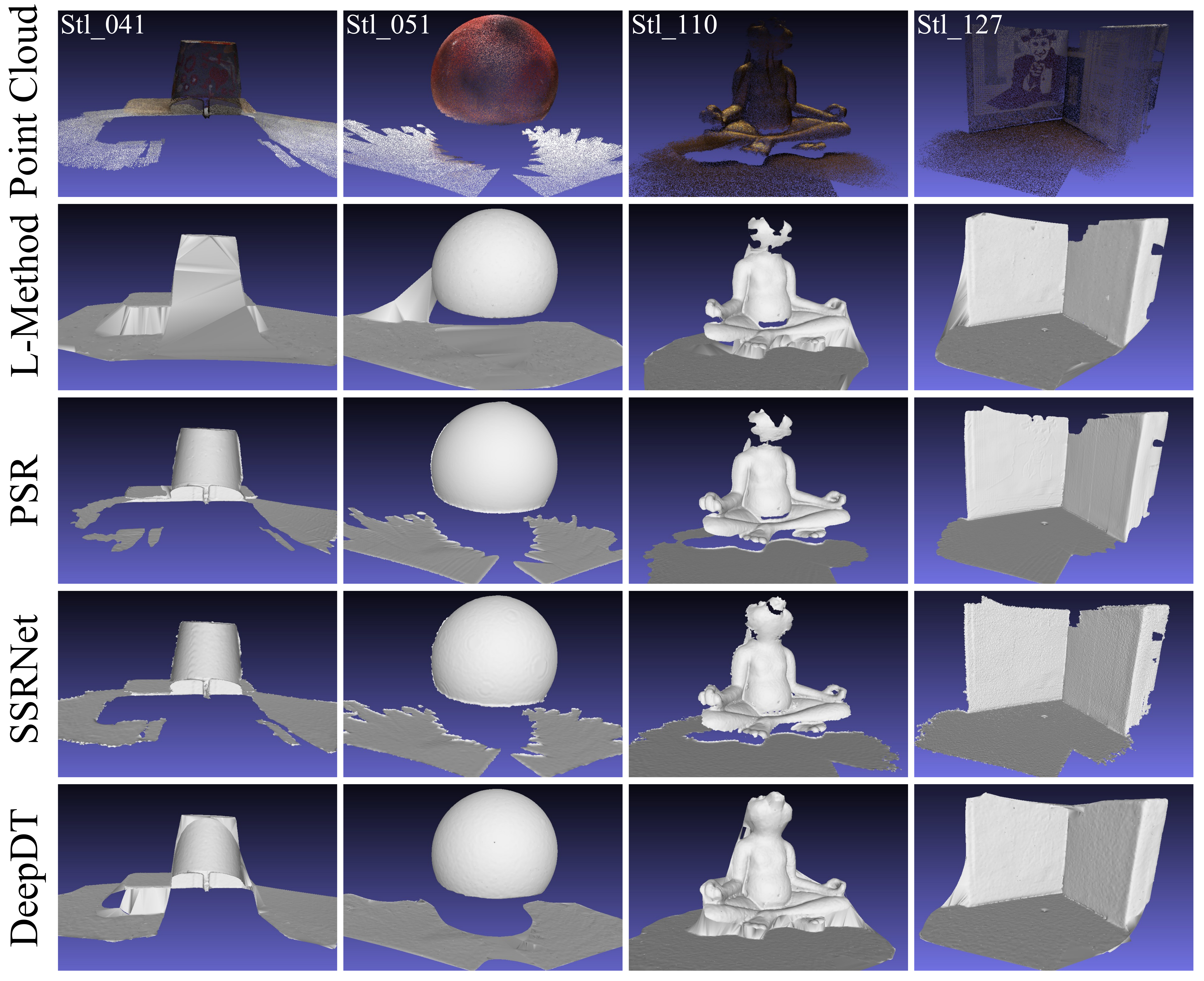}
	\end{center}
	\caption{Qualitative results on DTU test scans}
	\label{fig:test2}
\end{figure}

\subsection{Results on 3D Scans of Larger Scales}\label{sec:comparepsr}
In this section, we train and test DeepDT on DTU to verify the performance of our method on large 3D scan dataset. For each scan, we still use PSR (octree depth=10) with trimming value 8 to get labels of reference locations. We focus on whether the network can still maintain good reconstruction performance without using a sufficiently dense input. We do not use the entire point cloud, just a sample of two hundred thousand. Here we also cite traditional surface reconstruction method ( L-Method) and evaluate its performance on the same data. 
Although noise is inevitable in large 3D scanning datasets, we still apply Gaussian noise with zero mean and standard deviation 0.001 to the point clouds in DTU. Since L-Method needs to obtain visible sight information from the process of generating a point cloud from the depth map fusion, we apply the same noise to the depth map that it uses.

 \begin{table}
 	\centering
 	{
 		\begin{tabular}{lcccc}
 			\toprule
 			Metric\textbackslash{}Method & L-Method & PSR & SSRNet  & Ours\\
 			\midrule
 			DC-Mean &0.38  &0.35  &0.30  &0.37  \\
 			DC-Var. &0.60  &0.44  &0.09  &0.15  \\
 			\midrule
 			CD-Mean &1.21  &1.17  &1.46  &0.68  \\
 			\midrule
 	\end{tabular}}
 	\caption{Quantitative results on DTU. The lower the DTU Completeness (DC), the better. The Chamfer Distances (CD, lower is better) are in units of $ 10^{-8} $.}
 	\label{tab:DTU_results}
 \end{table}

Figure \ref{fig:test2} demonstrates that the two traditional methods  (PSR and L-Method) and the learning-based method (SSRNet) all generally preserve the geometry structure and complete the task well. 
Compared with L-Method, our method can avoid generating too many large triangular patches (Stl\_051), which should not be presented in the final surface and should be removed. Furthermore, one notable detail is that DeepDT can adaptively fill holes that need to be filled to a certain extent, rather than overfilling or remaining holes (Stl\_127, Stl\_110). 
Especially for open scene reconstruction tasks, our method can well avoid the phenomenon that some strange handles that should have been removed are not removed due to the wrong labeling of the tetrahedrons inside the surface. It can be seen in the first column of Figure \ref{fig:test2} that L-method cannot handle the inner surface of an open scene well, leaving many triangular patches that close the open scene to be removed. However, DeepDT reconstructs a smooth inner surface and reproduces the original shape of the open scene well. This greatly optimizes the visual effects of the final surface and validates our design of DeepDT.
 
Table \ref{tab:DTU_results} shows that DeepDT gets the competitive results. It is worth mentioning that the results of PSR and SSRNet are performed on the original point cloud without sampling or adding noise (We obtain PSR and SSRNet results from SSRNet paper). 
This means that DeepDT maintains good performance without using dense input, which benefits the method with the opportunity to explore better efficiency.

\begin{table*}
	\centering
	\small
	{
		\begin{tabular}{ccccccc}
			\toprule
			\multirow{2}{*}{Data}  & \multicolumn{3}{c}{CD Mean} & \multicolumn{3}{c}{CD RMS}                  \\
			\cmidrule(r){2-4} \cmidrule(r){5-7} 
			& ONet & SSRNet-S & DeepDT-S  & ONet & SSRNet-S & DeepDT-S \\
			\midrule
			Armadillo  & 93.46 & 0.028 & \textbf{0.021} & 168.59 & 0.131 & \textbf{0.028} \\
			Bunny      & 94.88 & 0.064 & \textbf{0.038} & 165.44 & 0.134 & \textbf{0.095} \\
			Dragon     & 40.69 & 0.053 & \textbf{0.029} & 74.99 & 0.208 & \textbf{0.081} \\
			\bottomrule
	\end{tabular}}
	\caption{Generalization performance test on Stanford 3D. The Chamfer Distances (CD, lower is better)
		are in units of $ 10^{-6} $. }
	\label{tab:generalization_on_stanford}

\end{table*}

\begin{figure}
	\begin{center}
		\includegraphics[width=\columnwidth]{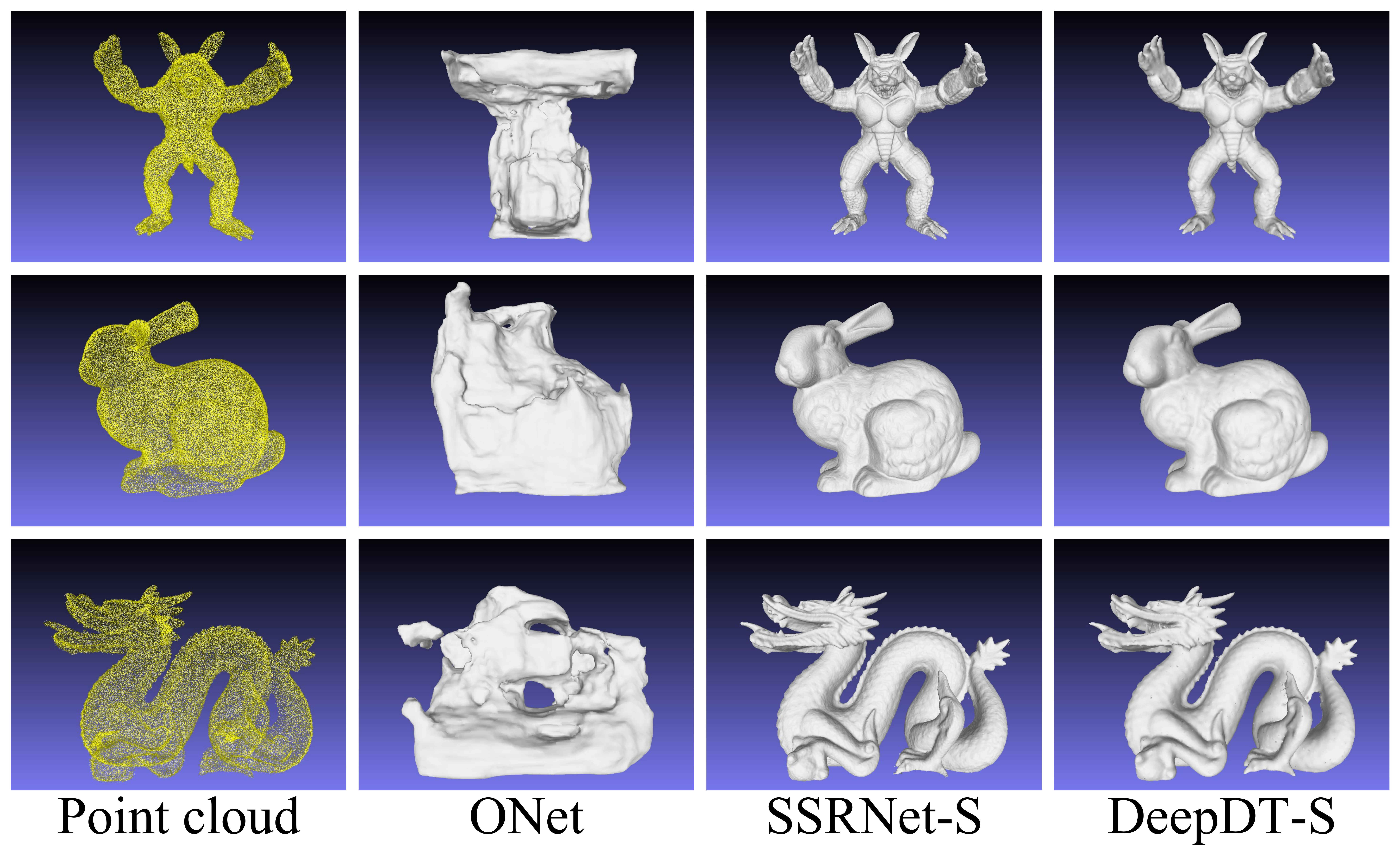}
	\end{center}
	\caption{Qualitative performance of ONet, SSRNet-S and DeepDT-S on Stanford 3D.}
	\label{fig:test5}
\end{figure}

\subsection{Efficiency}	
We test the efficiency of DeepDT on 1 GeForce RTX 2080 Ti GPU in
an Intel Xeon(R) CPU system 
with 40$ \times $2.20 GHz cores.
Here we choose DTU that better reflects the general efficiency. We also make a comparison with the learning-based method, SSRNet. 
We monitor the time consumption from inputting the point cloud to generating the final surface of each scan.
It takes an average of 8.985 seconds for DeepDT to complete a scan, and 200,000 points yield an average of 1,248,801 tetrahedrons.
SSRNet deals with the whole input with an average of 2.53 M points(1.20M vertices to be processed, which takes an average of 139.652 seconds). It can be accelerated by using more GPUs or reducing the input. However, the acceleration degree will not change linearly with the decrease of input points. This is mainly because SSRNet predicts the labels of vertices in an octree generated from the given point cloud. Even though the input data gets sparse, the spatial distribution of the point cloud does not change significantly, and the number of vertices will not decrease synchronously and linearly. Therefore, a sparse sample of the input will not significantly influence the efficiency of SSRNet.

\subsection{Generalization Capability}
For a learning-based method, it is important to have an outstanding
generalization ability, which reflects whether the
network can resist the threat from overfitting on a dataset.
Therefore, we perform experiments on the Stanford 3D dataset to verify whether DeepDT can complete the task well on a new dataset without retraining or changing network parameters.
We test DeepDT and SSRNet on Stanford 3D with the model trained on ShapeNet (DeepDT-S and SSRNet-S) to better assess the generalization ability. We still apply Gaussian noise with zero mean and standard deviation 0.001 to the point clouds in Stanford 3D. Here we also set the results of ONet as an evaluation reference. Quantitative and qualitative results are reported in Table \ref{tab:generalization_on_stanford} and Figure \ref{fig:test5}. The results of SSRNet are performed on the original dense point cloud without sampling or adding noise (We obtain SSRNet results from the paper of SSRNet). 
 
We can see that ONet fails to reconstruct the details and it cannot even restore the general shape. However, model SSRNet-S and DeepDT-S are able to better restore the geometric details of the surface without retraining. It is worth mentioning that the difference between the Stanford 3D dataset and the ShapeNet dataset is very large. This reflects the powerful generalization ability of SSRNet and DeepDT, making it possible to complete the reconstruction without changing the parameters or retraining on a 
new dataset.

\section{Discussion and Conclusion}
In this paper, we propose DeepDT, a novel learning-based method for surface reconstruction from 
Delaunay triangulation. The network integrates geometry features together with graph structural 
information for accurate labeling of tetrahedrons. In contrast to state-of-the-art traditional reconstruction method, L-Method, DeepDT needs no visibility information and outputs accurate meshes.
SSRNet adopts the strategy of dividing the input so that it can be adapted to sufficiently dense input to obtain better results. However, DeepDT is temporarily unable to use the same strategy to handle excessively large input. With its excellent performance, DeepDT just uses a sample of 200,000 points 
instead of the entire input 
to obtain competitive results in a more efficient way than SSRNet. Both DeepDT and SSRNet can complete the reconstruction of dense input well, but DeepDT is also good at dealing with the situation where there is not enough dense input. We have also noticed that when the input data is not dense enough, DeepDT witnessed a drop on texture finesse and clarity. An interesting direction could be to improve the visual effect of the DeepDT when maintaining good quantitative evaluation results.
We hope that our method will inspire more learning-based methods that no longer simply adapt the network to all the input data or reduce the input at the expense of accuracy to cope with 
reconstruction tasks, but focus on the quality, efficiency and 
generalization ability, making great contributions 
to daily applications.

\section*{Acknowledgements}
This work was supported by the National Natural Science Foundation of China under Grant 61772213, Grant 61991412 and Grant 91748204.
\bibliography{sample-bibliography}
\end{document}